\documentclass[runningheads]{llncs}
\usepackage{graphicx}
\usepackage{tikz}
\usetikzlibrary{positioning}
\usepackage{amsmath,amssymb}
\usepackage[procnames]{listings}
\usepackage{lstautogobble}
\usepackage{tabularx}
\usepackage{float}
\usepackage{paralist}
\usepackage{fancyvrb}
\usepackage{times}
\usepackage[colorlinks]{hyperref}
\usepackage{mathtools}
\usepackage{xspace}
\newcommand{\scenic}{{\sc Scenic}\xspace}
\newcommand{\verifai}{{\sc VerifAI}\xspace}
\newcommand{\hide}[1]{}

\newcounter{myctr}

\DeclareMathOperator*{\argmax}{arg\,max}

\newcolumntype{M}[1]{>{\centering\arraybackslash}m{#1}}
\renewenvironment{enumerate}[1]{\begin{compactenum}#1}{\end{compactenum}}

\begin{document}

\title{Parallel and Multi-Objective Falsification\\ with \scenic and \verifai}

\titlerunning{Parallel and Multi-Objective Falsification with \scenic and \verifai}

\author{Kesav Viswanadha\inst{1} \and
Edward Kim\inst{1} \and
Francis Indaheng\inst{1} \and
\\Daniel J. Fremont\inst{2} \and
Sanjit A. Seshia\inst{1}}

\authorrunning{K. Viswanadha et al.}

\institute{University of California, Berkeley \and
University of California, Santa Cruz}

\maketitle

\begin{abstract}
Falsification has emerged as an important tool for simulation-based verification of autonomous systems. In this paper, we present extensions to the \scenic scenario specification language and \verifai toolkit that improve the scalability of sampling-based falsification methods by using parallelism and extend falsification to multi-objective specifications. We first present a parallelized framework that is interfaced with both the simulation and sampling capabilities of \scenic and the falsification capabilities of \verifai, reducing the execution time bottleneck inherently present in simulation-based testing. We then present an extension of \verifai's falsification algorithms to support multi-objective optimization during sampling, using the concept of rulebooks to specify a preference ordering over multiple metrics that can be used to guide the counterexample search process. Lastly, we evaluate the benefits of these extensions with a comprehensive set of benchmarks written in the \scenic language.

\keywords{Runtime Verification \and Formal Methods \and Falsification \and Cyber-Physical Systems \and Autonomous Systems \and Parallelization}
\end{abstract}

%===========================================================
\section{Introduction}

The growing adoption of autonomous and semi-autonomous systems such as self-driving vehicles brings with it pressing questions about ensuring their safety and reliability. In particular, the increasing use of artificial intelligence (AI) and machine learning (ML) components requires significant advances in formal methods, of which simulation-based formal analysis is a key ingredient~\cite{verified_ai}.

Even with notable development in simulators and methods for simulation-based verification, there are four practical issues which require further advances in tools. First, simulation time can be a huge bottleneck, as falsification is typically done with high-quality, realistic simulators such as CARLA~\cite{carla}, which can be computation-intensive. Second, modeling interactive, multi-agent behaviors using general programming languages like Python can be very time-consuming. Third, autonomous systems usually need to satisfy multiple properties and metrics, with differing priorities, and convenient notation is needed to formally specify these. Fourth, we need to develop specification and sampling methods for falsification that can support multiple objectives.\par

There has been prior work that addresses these four issues separately, including several ideas for falsification or, conversely, optimization of an autonomous system subject to multiple objectives \cite{multi_objective_planning,multi_obj_cps,multi_obj_cps2,multi_obj_cps3}. There are other tools that address simulation-based testing of systems~\cite{staliro,falsification_rl,falsification_mtl}. In recent years, there have been significant advances in adversarial machine learning \cite{adversarial_ml,aml_av} as a way to find flaws with neural network-based systems. However, these methods tend to either focus on testing specific components of autonomous systems or are rather complex to use. More importantly, to our knowledge, no prior work has \textit{jointly} addressed all of these issues and demonstrated these in a single tool. 
In this paper, we do so by extending the open-source \verifai toolkit \cite{verifai}.\footnote{Documentation of the extensions covered in this paper is available at:\\ \href{https://verifai.readthedocs.io/en/kesav-v-multi-objective/}{https://verifai.readthedocs.io/en/kesav-v-multi-objective/}.} This tool is reasonably mature, having been demonstrated in multiple industrial case studies \cite{itsc20,cav20}. Our contributions to the toolkit support:
\begin{enumerate}
    \item {\it Parallelized falsification}, running multiple simulations in parallel;
    \item Falsification using the latest version of the {\it \scenic formal scenario specification language}, extending support to the ``dynamic'' features of \scenic for modeling interactive behaviors~\cite{scenic};
    \item The ability to specify for falsification {\it multiple objectives with priority orderings};
    \item A \textit{multi-armed bandit} algorithm that supports multi-objective falsification, and 
    \item Evaluation of these extensions with a comprehensive set of self-driving scenarios.
\end{enumerate}

These contributions have had a profound impact on the capabilities of \verifai. With parallel falsification, we were able to cut down drastically on execution time, achieving up to 5x speedup over the current falsification methods in \verifai using 5 parallel simulation processes. Using the multi-objective multi-armed bandit sampler, we were able to find scenarios which falsify five objectives at the same time.\par

%===========================================================
\section{Background}
\label{sec:background}

\scenic is a probabilistic programming language \cite{scenic_pldi,scenic,scenic_github} that allows users to intuitively model \textit{probabilistic scenarios} for multi-agent systems. 
A \textit{concrete scenario} is a set of objects and agents, together with values for their static attributes, initial state, and parameters of dynamic behavioral models describing how their attributes evolve over time. In other words, a concrete scenario defines a specific trace.
The state of each object or agent, such as a car, includes its semantic properties such as its position, orientation, velocity, color, model, etc.
We refer to the vector of such semantic properties as a \textit{semantic feature vector};
the concatenation of the semantic feature vectors of all objects and agents at a given time
instant defines the overall semantic feature vector at that time.
Agents also have behaviors defining a (possibly stochastic) sequence of actions for them to take as a function of the state of the simulation at each time step.
A \scenic program defines a \textit{distribution over concrete scenarios}: by sampling an initial state and then executing the behaviors in a simulator, many different simulations can be obtained from a single \scenic program.
\scenic provides a general formalism to express probabilistic scenarios for multiple domains, including traffic and other scenarios for autonomous vehicles, which can then be executed in a number of simulators including CARLA~\cite{carla}. In previous work on \verifai~\cite{verifai}, the tool supported an earlier version of \scenic without interactive, behavioral specifications. In this paper, we provide full support for \scenic's newer dynamic features.

\verifai is a Python toolkit that provides capabilities for verification of AI-based systems \cite{verifai_github}. A primary capability is \textit{falsification}, the systematic search for inputs to a system that falsify a specification given in temporal logic or as a cost function. \verifai can use \scenic as an environment modeling language, sampling from the distribution over semantic feature vectors defined by a \scenic program to generate test cases.
It then simulates these cases, obtaining trajectories.

After simulating a test case, \verifai evaluates the system's specification over the obtained trajectory, saving the results for offline analysis.
These results are also used to guide further falsification, specifically by \verifai's \textit{active samplers}, such as the cross-entropy sampler~\cite{cross-entropy}.
These samplers use the history of previously generated samples and their outcomes in simulation to drive the search process to find more counterexamples.

%==================================================================================
\section{Parallel Falsification}
\label{sec:parallel-falsif}

In the typical pipeline used by a \verifai falsifier driven by a \scenic program, semantic feature vectors (parameters) are generated using samplers in either \scenic or \verifai. These parameter values are then sent by the \verifai server to the client simulator to configure a simulation and generate a corresponding trajectory. This trajectory is then evaluated by the monitor, deemed either a safe example or a counterexample, and added to the corresponding table in the falsifier. Naturally, a bottleneck of this process is the generation of the trajectory in the simulator, as this is a rather compute-intensive task that can take a minute or more per sample, depending on the scenario description.\par

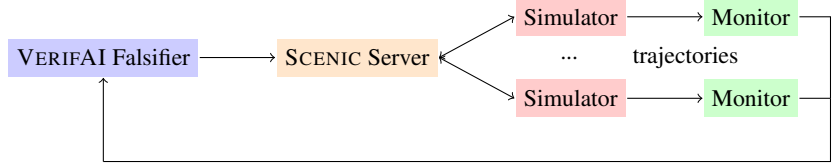
\begin{figure}
    \centering
    \begin{tikzpicture}[
    verifainode/.style={rectangle, fill=blue!20, very thick, minimum size=5mm},
    servernode/.style={rectangle, fill=orange!20, very thick, minimum size=5mm},
    monitornode/.style={rectangle, fill=green!20, very thick, minimum size=5mm},
    simnode/.style={rectangle, fill=red!20, very thick, minimum size=5mm},
    ]
    \node[verifainode]      (falsifier)                              {\verifai Falsifier};
    \node[servernode]      (scenicserver)       [right=of falsifier] {\scenic Server};
    \node[simnode]         (sim1)       [above right=0mm and 1cm of scenicserver] {Simulator};
    \node[simnode]         (sim2)       [below right=0mm and 1cm of scenicserver] {Simulator};
    \node[monitornode]      (monitor1)           [right=of sim1] {Monitor};
    \node[monitornode]      (monitor2)           [right=of sim2] {Monitor};
    \node                  (simright)   [right=3mm of monitor2]{};
    \node                  (simdots)    [right=1.5cm of scenicserver]{...};
    \node                  (traj)        [right=5mm of simdots]{trajectories};
    \node                  (fals) [below=1cm of falsifier]{};
    \draw[->] (falsifier.east) -- (scenicserver.west);
    \draw[-] (monitor1.east) -| (simright.center);
    \draw[-] (monitor2.east) -- (simright.center);
    \draw[->] (sim1.east) -- (monitor1.west);
    \draw[->] (sim2.east) -- (monitor2.west);
    \draw[<->] (scenicserver.east) -- (sim1.west);
    \draw[<->] (scenicserver.east) -- (sim2.west);
    \draw[-] (simright.center) |- (fals.center);
    \draw[->] (fals.center) -- (falsifier.south);
    
    \end{tikzpicture}
\caption{Parallelized pipeline for falsification using \verifai.}
\label{fig:parallel_falsifier_pipeline}
\end{figure}

We present an improvement on this pipeline by parallelizing it using the Python library Ray~\cite{ray}, which encapsulates process-level parallelism optimized for distributed execution of computation-intensive tasks. Fig.~\ref{fig:parallel_falsifier_pipeline} illustrates the new setup: we instantiate multiple instances of the simulator and open multiple \scenic server connections from \verifai to the simulator instances for performing simulations (the connections now being bidirectional so that the behavior models in the \scenic program can respond to the current state of the simulation). We then aggregate the results of these simulations into a single error table documenting all the counterexamples found during falsification.

\section{Multi-Objective Falsification}
There are typically many different metrics of interest for evaluating autonomous systems. For example, there are many well-known metrics used in the autonomous driving community to measure safety: no collisions, obeying traffic laws, and maintaining a minimum safe distance from other objects, among others~\cite{metrics}. It is also natural to assert, for example, that it is more important to avoid collisions than to follow traffic laws. We now discuss how to specify these metrics and their relative \emph{priorities}.

\subsection{Specification of Multiple Objectives Using Rulebooks}

Let $\rho(x)$ be a function mapping a simulation trajectory generated by \scenic or \verifai to a vector-valued objective, where $\rho_j(x)$ is defined as the value of the $j$\nobreakdash-th metric. Censi et al.~\cite{rulebooks} have developed a way to specify preferences over these metrics using a \textit{rulebook} denoted by $\mathcal{R}$ -- a directed acyclic graph (DAG) where the nodes are the metrics and a directed edge from node $i$ to node $j$ means $\rho_i(x)$ is more important than $\rho_j(x)$. We denote this using the $>_R$ operator, e.g. $\rho_i >_R \rho_j$.\par
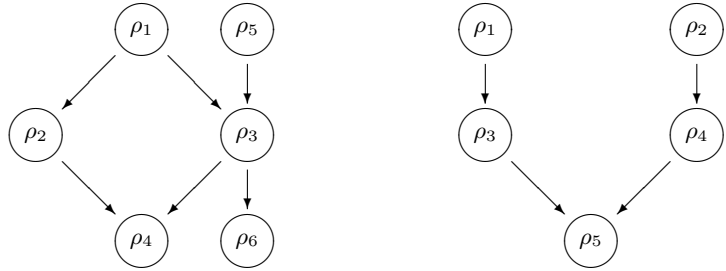
\begin{figure}[tb]
\centering
\begin{picture}(300,90)
  \put(30,70){\makebox(20,20){$\rho_1$}}
  \put(40,80){\circle{20}}
  \put(70,30){\makebox(20,20){$\rho_3$}}
  \put(80,40){\circle{20}}
  \put(30,-10){\makebox(20,20){$\rho_4$}}
  \put(40,0){\circle{20}}
  \put(-10,30){\makebox(20,20){$\rho_2$}}
  \put(0,40){\circle{20}}
  \put(70,70){\makebox(20,20){$\rho_5$}}
  \put(80,80){\circle{20}}
  \put(70,-10){\makebox(20,20){$\rho_6$}}
  \put(80,0){\circle{20}}
  \put(50,70){\vector(1,-1){20}}
  \put(30,70){\vector(-1,-1){20}}
  \put(10,30){\vector(1,-1){20}}
  \put(70,30){\vector(-1,-1){20}}
  \put(80,67){\vector(0,-1){15}}
  \put(80,27){\vector(0,-1){15}}
  \put(160,70){\makebox(20,20){$\rho_1$}}
  \put(170,80){\circle{20}}
  \put(240,30){\makebox(20,20){$\rho_4$}}
  \put(250,40){\circle{20}}
  \put(200,-10){\makebox(20,20){$\rho_5$}}
  \put(210,0){\circle{20}}
  \put(160,30){\makebox(20,20){$\rho_3$}}
  \put(170,40){\circle{20}}
  \put(240,70){\makebox(20,20){$\rho_2$}}
  \put(250,80){\circle{20}}
  \put(170,67){\vector(0,-1){15}}
  \put(180,30){\vector(1,-1){20}}
  \put(240,30){\vector(-1,-1){20}}
  \put(250,67){\vector(0,-1){15}}
\end{picture}
\caption{Left: example rulebook over functions $\rho_1 \dots \rho_6$ \cite{rulebooks}. Right: graph $G$ used in experiments.}
\label{fig:rulebook_example}
\end{figure}
Fig.~\ref{fig:rulebook_example} shows an example of a rulebook over six metrics $\rho_1, \dots, \rho_6$. In this example, we can make several inferences, such as $\rho_1$ is more important than $\rho_3$, $\rho_3$ is more important than $\rho_4$, and $\rho_5$ is more important than $\rho_3$. However, there are also many pairs of objective components that cannot be compared; for example $\rho_1$ and $\rho_5$. We would like to have a way to order objective vectors to know which values are maximally violating of the specification during active sampling. Because of these indeterminate incomparisons, the rulebook $\mathcal{R}$ only allows for a \textit{partial ordering} $\succ$ over the objective vectors. Intuitively, we can think of this partial ordering as preferring examples that have lower values of higher priority objectives since we are trying to minimize the values of each objective for falsification. However, if there is any other indeterminate or higher priority objective that has a higher value, the $\succ$ relation does not hold. To satisfy these properties, we define our $\succ$ operator as follows:
$$\rho(x_1) \succ \rho(x_2) \triangleq \forall i \, \bigl(\rho_i(x_2) < \rho_i(x_1) \implies \exists j \neq i \, ( \rho_j >_R \rho_i \land \rho_j(x_1) < \rho_j(x_2) ) \bigr)$$
As an example, consider our rulebook from Fig.~\ref{fig:rulebook_example}. Let $\rho(x_1) = \begin{bmatrix} 1 & 1 & 1 & 1 & 1 & 1 \end{bmatrix}^T$, and $\rho(x_2) = \begin{bmatrix} 1 & 1 & 2 & 1 & 0 & 1 \end{bmatrix}^T$. In this case we have $\rho(x_2) \succ \rho(x_1)$ because $\rho_5(x_2) < \rho_5(x_1)$, and even though $\rho_3(x_2) > \rho_3(x_1)$, $\rho_5 >_R \rho_3$ according to the rulebook, so the comparison of $\rho_5$ for the trajectories takes precedence. Since the rulebook defines a partial ordering over values of $\rho$, it is possible to have two trajectories $x_1$ and $x_2$ such that $\rho(x_1) \not \succ \rho(x_2)$ and $\rho(x_2) \not \succ \rho(x_1)$. In such cases, both values of $\rho$ are maintained in the sampling algorithm; see below for more details.

\subsection{Multi-Objective Active Sampling}

When performing active sampling to search for unsafe test inputs, we need a specialized sampler to support having multiple objectives to guide the search process. Most of the samplers previously available in \verifai focused either entirely on exploration of the search space or entirely on exploitation to find unsafe inputs; we present a sampler that balances these and builds up increasingly-violating counterexamples in the multi-objective case.

\noindent \textbf{The Multi-Armed Bandit Sampler.}
We present a more robust version of \verifai's cross-entropy sampler called the \textit{multi-armed bandit sampler}; the idea of this sampler is to balance the trade-off between exploitation and exploration. To understand the motivation for the sampler, we first look at the formulation of the multi-armed bandit problem. Consider a bandit which has multiple lotteries, or ``arms", to choose from, each being a random variable offering a probabilistic reward. The bandit does not know ahead of time which arm gives the highest expected reward, and must learn this information by efficiently sampling various arms, while also maximizing average earned reward during the sampling process.

Carpentier et al.~\cite{ucb} present the Upper Confidence Bound (UCB) Algorithm that effectively balances both of these goals, subject to a confidence parameter $\delta$, by sampling the arm $j$ that minimizes a quantity $Q_j$ dependent on the number of timesteps $t$, the number of times the arm $j$ was sampled $T_j(t - 1)$, the observed reward of arm $j$ given by $\hat{\mu}_j$, and the confidence parameter $\delta$:
$$Q_j = \hat{\mu}_j + \sqrt{\frac{2}{T_j(t - 1)}\ln\left(\frac{1}{\delta}\right)}$$
Qualitatively, this works as a balance between exploitation of the reward distribution learned so far (the first term), and exploration of seldom-sampled arms (the second term). We can easily see that this can be readily adapted to our cross-entropy sampler in \verifai, which splits the range of each sampled variable into $N$ equally spaced \textit{buckets}, which can be considered the ``arms". We take $\hat{\mu}_j$ to be the proportion of counterexamples found in bucket $j$.\par
To compute $\mu_j$ for a vector-valued objective, we present the following incremental algorithm which builds up counterexamples that falsify more and more objectives (according to the priority order) over time. The steps of this algorithm are as follows. This assumes that the sampler is responsible for generating a $d$-dimensional feature vector.\\
\textbf{Setup}
\begin{enumerate}
    \item Split the range of each component of the feature vector into $N$ buckets, as in the cross-entropy sampler.
    \item Initialize matrix $T$ of size $d \times N$ where $T_{ij}$ will keep track of the number of times that bucket $j$ was visited for variable $x_i$.
    \item Initialize a dictionary $c$ mapping each maximal counterexample found so far to a matrix $c_b$ of size $d \times N$ where $c_{b,ij}$ counts how many times sampling bucket $j$ for variable $x_i$ resulted in the specific counterexample $b$.
    \item Sample from each bucket once initially, updating $c$ and $T$ according to the update algorithm described below. The purpose of this is to avoid division by zero when computing $Q$, as $T_j(t - 1) = 0$ at initialization \cite{bandit_algs}.
\end{enumerate}
\textbf{Sampling}
\begin{enumerate}
    \item Compute a matrix $\hat{\mu}$ where $\hat{\mu}_{ij}$ represents the observed reward from sampling bucket $j$ for variable $i$ by taking $\sum_b c_{b,ij}$.
    \item Compute a matrix $Q$ based on the upper confidence bound formula above. For the confidence parameter, we use a time-dependent value of $\frac{1}{\delta} = t$.
    \item To sample $x_i$, take the bucket $j^* = \argmax_j Q_{ij}$. \textit{Break ties uniformly at random.} This is a key step in the sampling process as it is frequently the case initially that several buckets will have the exact same $Q_j$ value, so we need to avoid bias towards any specific bucket. Sample uniform randomly within the range represented by bucket $j^*$.
\end{enumerate}
\textbf{Updating Internal State}
\begin{enumerate}
    \item Given the objective vector value $\rho$, we compute our vector of booleans $b$ as described above.
    \item If $b$ does not exist in the dictionary $c$ and is among the set of maximal counterexamples found so far, i.e. $\forall b' \in c, b' \not \succ b$ as defined by the rulebook $\mathcal{R}$, add $b$ as a key to the dictionary $c$ and initialize its value as $0^{d \times N}$.
    \item For any $b' \in c$ such that $b \succ b'$, remove $b'$ from $c$.
    \item Increment the count $c_b$ at each position $c_{b,ij}$ for the bucket $j$ sampled from $x_i$.
\end{enumerate}

\section{Evaluation}

We present a set of experiments designed to evaluate (i) the speedup in simulation time that we expect to see from parallelization; (ii) the benefits of the multi-armed bandit sampler in balancing exploration and exploitation; and (iii) the improved capabilities of falsification to support multiple objectives. We have developed a library of \scenic scripts\footnote{Full listing and source code of these \scenic scripts is available at:\\ \href{https://github.com/findaheng/Scenic/tree/behavior\_prediction/examples/carla/Behavior\_Prediction}{https://github.com/findaheng/Scenic/tree/behavior\_prediction/examples/carla/Behavior\_Prediction}.} based on the list of pre-crash scenarios described by the National Highway Traffic Safety Administration (NHTSA) \cite{nhtsa}. For a list of the scenarios, see the \hyperref[sec:appendix]{Appendix}. These scripts cover a wide variety of common driving situations, such as driving through intersections, bypassing vehicles, and accounting for pedestrians.

We selected 7 of these scenarios, running the \verifai falsifier on each one in CARLA~\cite{carla} for 30 minutes, with individual simulations limited to 300 timesteps ($\sim$30 seconds). For all of these scenarios, the monitor specifies that the centers of the ego vehicle and other vehicles must stay at least 5 meters apart at all times. This specification means that counterexamples approximately correspond to collisions or near-collisions. All parallelized experiments were run using 5 worker processes to perform simulation.\par

\begin{figure}
    \centering
    \includegraphics[width=0.7\textwidth]{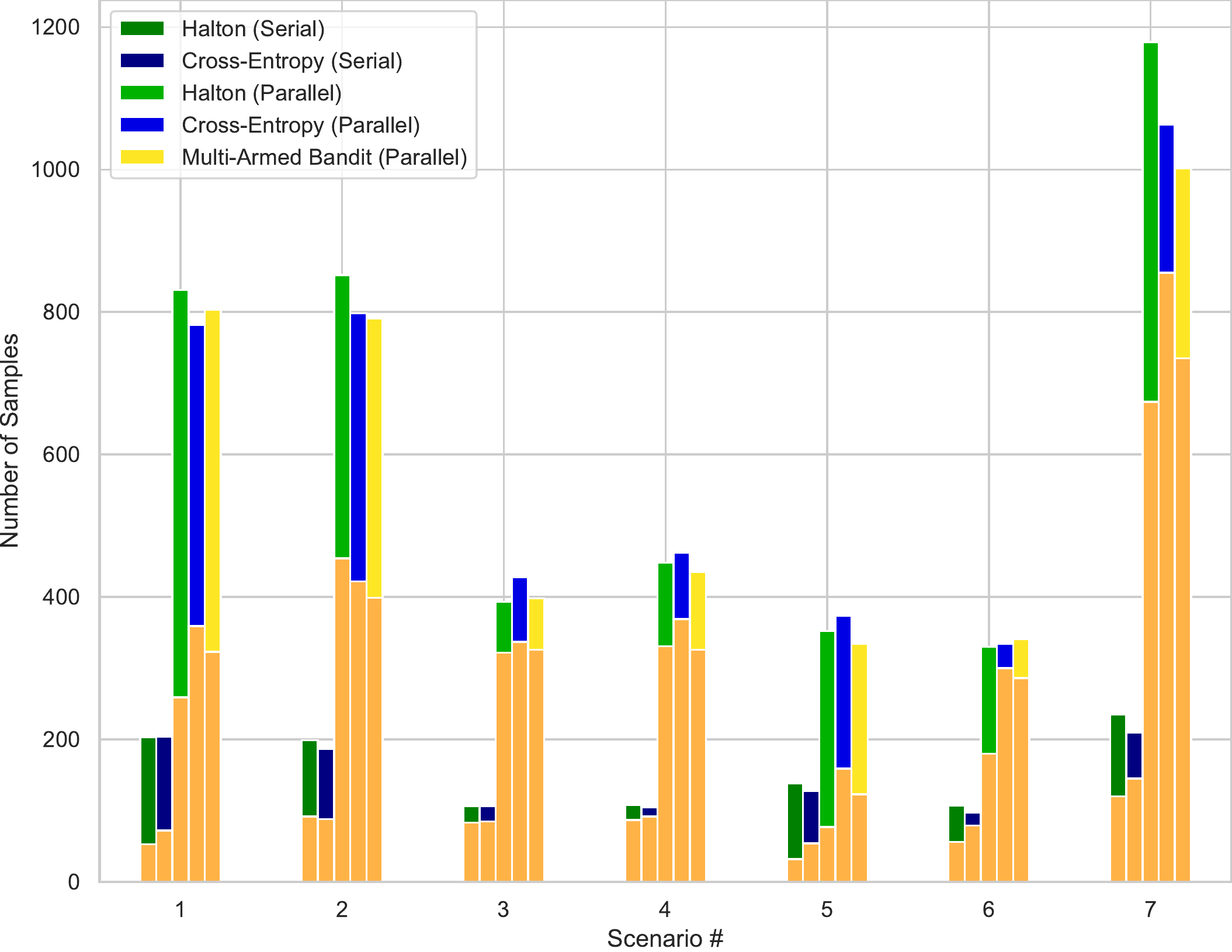}
    \caption{Comparison of (i) the serial and parallel versions of the falsifier for cross-entropy and Halton sampling and (ii) the multi-armed bandit sampler with the cross-entropy and Halton samplers all in parallel. The orange part of the bars represent the number of counterexamples found out of the total number of samples.}
    \label{fig:sampler_comparison}
\end{figure}

Fig.~\ref{fig:sampler_comparison} shows the results of running these scenarios with a variety of configurations. First, across the scenarios, we observed a 3-5x speedup in the number of simulations using 5 parallel simulation processes. The variation in the number of samples generated can be attributed to \textit{termination conditions} set in \scenic, which terminate simulations early if specific conditions are met. For some of these scenarios, termination occurred much sooner on average than other scenarios, leading to more simulations finishing in 30 minutes. These values also serve as partial evidence of the effectiveness of the multi-armed bandit sampler compared to cross-entropy, as the proportion of counterexamples found is comparable for the two samplers despite the increased exploration component in the multi-armed bandit sampler.\par
To validate the scalability and explorative aspect of parallelized falsification, we present two metrics in Table~\ref{tab:metrics}. The first metric is the \textit{speedup factor}, which is the ratio of the number of sampled scenarios in parallel versus serial falsification, averaged across the Halton and cross-entropy samplers. We are also interested in a metric of coverage of the scenario search space, as this ensures that a wide range of scenarios are tested by falsification. To this end, we present the \textit{confidence interval width ratio} metric. This metric is computed by generating a 95\% confidence interval \cite{clopper_pearson} which provides a lower and upper bound on the probability that a randomly generated scenario results in unsafe behavior. Since confidence intervals are generated with the assumption of uniform random sampling, we only compute them for the serial and parallel Halton samplers since they are an approximation of random sampling. We take the ratio of the widths of the intervals in the parallel versus serial case to compare how tight we are able to make the bound in each case with the same level of confidence. The width of the interval in the parallel case is significantly smaller - up to half the width of the serial case. Since the width of the interval is proportional to $1/\sqrt{n}$ for $n$ samples, this makes intuitive sense and can be viewed as having double the coverage of the search space.
\begin{table}[ht]
    \centering
    \caption{The speedup factor and confidence interval width ratio metrics for the 7 scenarios.}
    \begin{tabular}{M{0.25\linewidth}|M{0.09\linewidth}M{0.09\linewidth}M{0.09\linewidth}M{0.09\linewidth}M{0.09\linewidth}M{0.09\linewidth}M{0.09\linewidth}}
        \hline Scenario \# & 1 & 2 & 3 & 4 & 5 & 6 & 7 \\
        \hline \textbf{Speedup Factor} & 3.96 & 4.27 & 3.87 & 4.27 & 2.73 & 3.26 & 5.04 \\
        \textbf{CI Width Ratio} & 0.51 & 0.48 & 0.48 & 0.53 & 0.61 & 0.56 & 0.44 \\
    \end{tabular}
    \label{tab:metrics}
\end{table}

\begin{figure}
    \centering
    \includegraphics[width=0.98\textwidth]{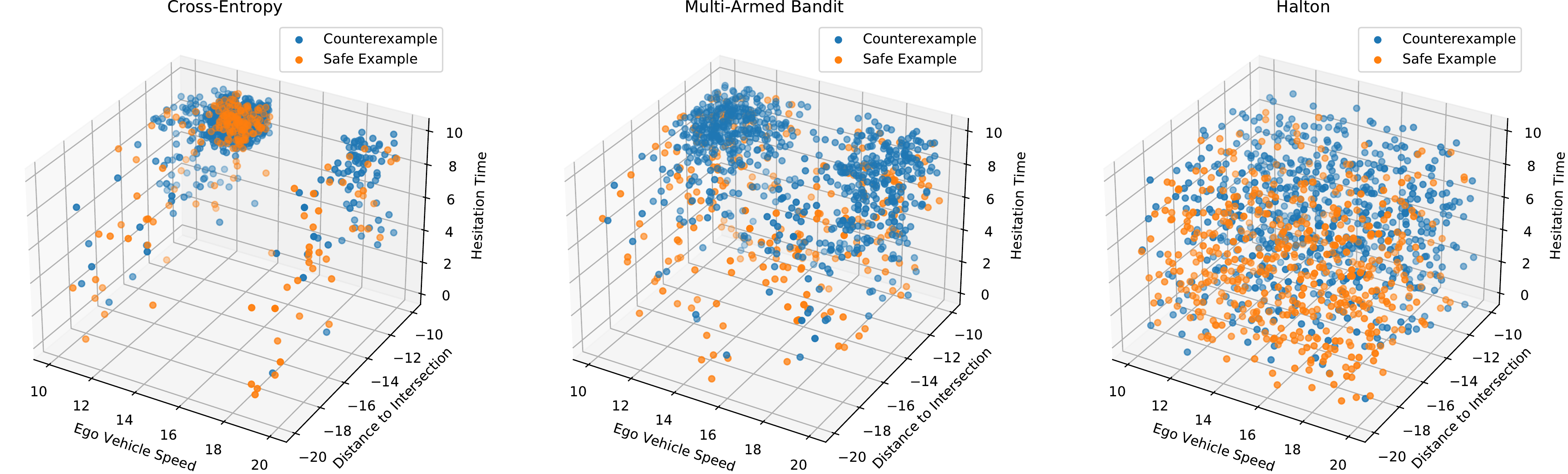}
    \caption{Comparison of points sampled for cross-entropy, MAB, and Halton samplers.}
    \label{fig:sampler_plots}
\end{figure}

Figs.~\ref{fig:sampler_comparison} and \ref{fig:sampler_plots} show the qualitative benefits of the multi-armed bandit sampler. The number of counterexamples generated by the multi-armed bandit sampler is higher than for the Halton sampler, but only slightly lower than cross-entropy. However, we can clearly see that multi-armed bandit sampling achieves a balance between number of counterexamples and their diversity that cross-entropy and Halton do not.\par
To demonstrate the effectiveness of the multi-objective multi-armed bandit sampler in falsifying multiple objectives, we used a \scenic program that instantiates the ego vehicle, along with $m$ adversarial vehicles at random positions with respect to a 4-way intersection and has all of them drive towards the intersection and either go straight or make a turn. The monitor, similarly to before, specifies metric components $\rho_j$ which say the ego vehicle must stay at least 5 meters away from vehicle $j$. We use the following three rulebooks: a completely disconnected graph representing no preference ordering, a linked list structure $L \triangleq \rho_1 >_R \rho_2 >_R ... >_R \rho_5$ representing a total ordering, and the graph $G$ on the right in Fig.~\ref{fig:rulebook_example}. We found that when using $L$ or $G$, we were able to falsify 4 of the 5 objectives with serial falsification, and all 5 objectives in the parallel case. When having no preference ordering, we were able to falsify 3 of the 5 objectives with serial falsification and 4 of the 5 objectives in the parallel case. By contrast, when we combined all of these objectives in disjunction as one single objective (such that only falsifying all 5 objectives is considered unsafe), the cross-entropy sampler was unable to find any counterexamples.\par

\section{Conclusion and Future Work}

The extensions to \scenic and \verifai we report in this paper address important problems in simulation-based falsification. First, we cut down significantly on execution time by supporting parallel simulations. Second, we allow the simple specification of high-level yet complex scenarios using the interface between dynamic \scenic and \verifai. Third, we support multi-objective specification through the formalism of rulebooks. Lastly, we are able to falsify these multi-objective specifications in a way that is intuitive and scalable using the multi-armed bandit sampler. We hope these extensions prove useful to developers of autonomous systems.\par
There are many directions for future work. For example, it might be interesting to see if generating random topological sorts of the rulebooks to create total ordering works well in practice. One could also run covariance analysis on the features to determine if they can be jointly optimized for better active sampling. Lastly, there has been some work in connecting these ideas to real-world testing \cite{itsc20}, but especially with multi-objective falsification, this is an interesting future direction.

\newpage
\bibliographystyle{splncs04}
\bibliography{references}
\newpage
\clearpage\section*{Appendix}
\label{sec:appendix}
\subsection*{List of Evaluation Scenarios}
\begin{table}[]
    \centering
    \begin{tabular}{M{0.12\linewidth}|M{0.48\linewidth}|M{0.48\linewidth}}
        \hline Scenario \# & Scenario Description & Related NHTSA Pre-Crash Scenario(s) \cite{nhtsa} \\
        \hline 1 & The ego vehicle drives straight at a 4-way intersection and must suddenly stop to avoid collision when an adversary vehicle from an oncoming parallel lane makes an unprotected left turn. & Scenario 30: Left Turn Across Path, Opposite Direction \\
        \hline 2 & The ego vehicle makes an unprotected left turn at a 4-way intersection and must suddenly stop to avoid collision when an adversary vehicle from an oncoming parallel lane drives straight. & Scenario 30: Left Turn Across Path, Opposite Direction \\
        \hline 3 & The ego vehicle performs a lane change to bypass a leading vehicle before returning to its original lane. & Scenario 14: Changing Lanes, Same Direction \\
        \hline 4 & A trailing vehicle performs a lane change to bypass the ego vehicle before returning to its original lane. & Scenario 14: Changing Lanes, Same Direction \\
        \hline 5 & The ego vehicle performs a lane change to bypass a leading vehicle, but cannot return to its original lane because the leading vehicle accelerates. The ego vehicle must then slow down to avoid collision with the leading vehicle in its new lane. & Scenario 14: Changing Lanes, Same Direction \newline Scenario 20: Rear-End, Striking Maneuver \newline Scenario 22: Rear-End, Lead Vehicle Moving \\
        \hline 6 & The ego vehicle must suddenly stop to avoid collision when a pedestrian crosses the road unexpectedly. & Scenario 10: Pedestrian, No Maneuver \\
        \hline 7 & Both the ego vehicle and an adversary vehicle must suddenly stop to avoid collision when a pedestrian crosses the road unexpectedly. & Scenario 10: Pedestrian, No Maneuver \\
    \end{tabular}
\end{table}

\end{document}